\documentclass[5pt]{article}

\usepackage[letterpaper]{geometry}
\usepackage{spconf,amsmath,epsfig}
\usepackage{enumitem}
\usepackage{multirow}

\usepackage{fancyhdr}
\begin{document}\sloppy

% Example definitions.
% --------------------
\def\x{{\mathbf x}}
\def\L{{\cal L}}

% Title.
% ------
\title{Investigation of Different Skeleton Features for CNN-based 3D Action Recognition}
%
% Single address.
% ---------------
\name{Zewei Ding, Pichao Wang*\thanks{*Corresponding author}, Philip 
O. Ogunbona, Wanqing Li}
\address{Advanced Multimedia Research Lab, University of Wollongong, Australia \\
	\{zd027, pw212, philipo, wanqing\}@uow.edu.au}
%
% For example:
% ------------
%\address{School\\
%	Department\\
%	Address\\
%   Email}
%
% Two addresses (uncomment and modify for two-address case).
% ----------------------------------------------------------
%\twoauthors
%  {A. Author-one, B. Author-two\sthanks{Thanks to XYZ agency for funding.}}
%	{School A-B\\
%	Department A-B\\
%	Address A-B}
%  {C. Author-three, D. Author-four\sthanks{The fourth author performed the work
%	while at ...}}
%	{School C-D\\
%	Department C-D\\
%	Address C-D\\
%   Email}
%

\maketitle

\begin{abstract}
Deep learning techniques are being used in skeleton based action 
recognition tasks and outstanding performance has been reported. Compared with 
RNN based methods which tend to overemphasize temporal information, CNN-based 
approaches can jointly capture spatio-temporal information from texture color 
images encoded from skeleton sequences. There are several skeleton-based 
features that have proven effective in RNN-based and handcrafted-feature-based 
methods. However, it remains unknown whether they are suitable for CNN-based 
approaches. This paper proposes to encode five spatial skeleton features into 
images with different encoding methods. In addition, the performance 
implication of different joints used for feature extraction is studied. The 
proposed method achieved state-of-the-art performance on NTU RGB+D dataset for 
3D human action analysis. An accuracy of 75.32\% was achieved in Large Scale 3D 
Human Activity Analysis Challenge in Depth Videos.
\end{abstract}
\begin{keywords}
Skeleton, 3D Action recognition, Convolutional Neural Networks
\end{keywords}
\section{Introduction}
\label{sec:intro}
Recognition of human actions has recently attracted increased interest because 
of its applicability in systems such as human-computer interaction, game 
control, and intelligent surveillance. With the development of cost-effective 
sensors such as Microsoft Kinect cameras, RGB-D-based recognition has 
almost become 
commonplace~\cite{li2010action,Wang2015,pichaoTHMS,Pichaocvpr2017}. Among the 
three most common input streams (RGB, depth, and skeleton),  RGB is the most 
popular and widely studied. However, it suffers the challenge of 
pose ambiguity due to the loss of 3D information. On the other hand, depth and 
skeleton which capture 3D information of human bodies inherently overcome 
this challenge. 
	
Skeleton has the advantage of being invariant to viewpoints or 
appearances compared with depth, thus suffering less intra-class 
variance~\cite{Zhang2017}. Furthermore, learning over skeleton is simple 
because they are higher-level information based on advanced pose 
estimation. The foregoing observations motivated the study of skeleton-based 
human action recognition in this paper.
	
The methods based on handcrafted skeleton 
features~\cite{xia2012view,wang2014mining,Vemulapalli2014} have the drawback of 
dataset dependency while methods based on deep learning 
techniques have achieved outstanding performance. Currently, there are mainly 
two ways of using deep learning techniques to capture the spatio-temporal 
information in skeleton sequences; Recurrent Neural Neural Networks 
(RNNs) and Convolutional Neural Networks (CNNs). RNNs are adopted to capture 
temporal information from extracted spatial skeleton features. The performance 
relies much on the effectiveness of the extracted spatial skeleton features due 
to the sequential flow of information. Moreover, the temporal information can 
be easily overemphasized especially when the training data is insufficient, 
leading to over-fitting~\cite{Wang2016}. 
	
In contrast, CNNs directly extract information from texture images which 
are encoded from skeleton sequences. Wang et al~\cite{Wang2016} used Joint 
Trajectory Maps (JTM) to encode body joint trajectories (positions, motion 
directions, and motion magnitudes) of each time instance into HSV images. In the 
images, spatial information is represented by positions and the dynamics is 
represented by colors. Hou et al~\cite{Hou2016} adopted Skeleton Optical Spectra 
(SOS) to encode dynamic spatial-temporal information. Li et al~\cite{Li2017} 
adopted joint distances as spatial features and a colorbar was used for 
color-encoding. In the images, textures of rows capture spatial information and 
textures of columns capture temporal information. Currently, the spatial 
features used for encoding are relatively simple (joints positions and 
pair-wise distances). 
	
% %Following the CNN-based approach, this paper proposes to %encode 
% % spatial skeleton features into texture color images, based on which CNNs are 
% % used to recognize corresponding actions. 
		
\begin{figure*}[!t]
	\centering
	\includegraphics[width=\linewidth]{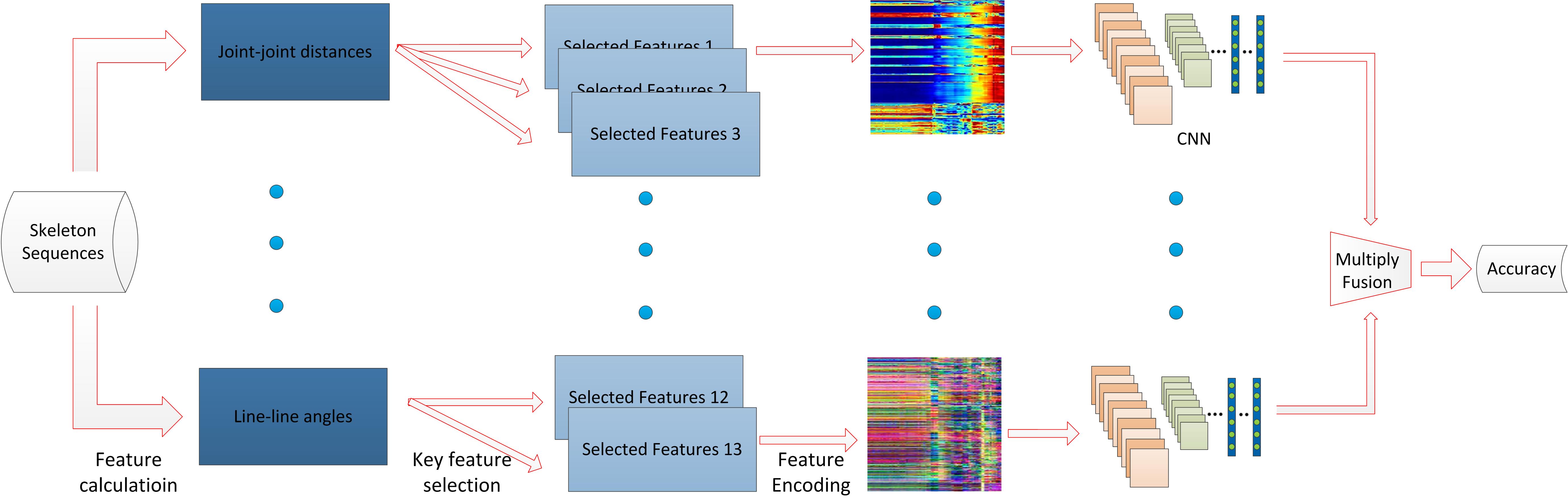}
	\caption{The framework of the proposed method}
	\label{fig:framework}
\end{figure*}
		
Following the CNN-based approach, this paper investigates encoding richer 
spatial features into texture color images, including features between two or 
more joints. Specifically, inspired by the work from Zhang et 
al\cite{Zhang2017}, the encoding of the following five types of spatial 
features is studied: joint-joint distances (JJd), joint-joint orientations 
(JJo), joint-joint vectors (JJv), joint-line distances (JLd), line-line angles 
(LLa). Each kind of feature is encoded into images in two or more ways to 
further explore the spatio-temporal information. CNN is adopted to train and 
recognize corresponding actions and score fusion is used to make a final 
classification. The effectiveness of this kind of approach has been verified 
in~\cite{Wang2016,Li2017,Wang2015}. The investigation is conducted on NTU RGB+D 
Dataset~\cite{Shahroudy2016} and achieves state-of-the-art performance.

The rest of the paper is organized as follows. Section~\ref{sec:method} 
introduces the proposed method and, in Section~\ref{sec:experiment}, 
experimental results and discussions are described. The conclusion and future 
work are presented in Section~\ref{sec:conclusion}.
	
% %Du et al\cite{Du2015} divided the skeleton into five parts and then 
% used a hierarchical 
% RNN consisting of five bidirectional 
% sub-RNNs to model the dynamics. 

\section{Proposed Method}
\label{sec:method}
As illustrated in Fig.~\ref{fig:framework}, the proposed method consists of five 
main components, namely spatial feature extraction from input skeleton 
sequences, key feature selection, texture color image encoding from key 
features, CNN model training based on images, and the score fusion. There are 
five types of features extracted from all joint combinations including JJd, 
JJo, JJv, JLd and LLa. Key features of certain joint combinations are then 
chosen for color encoding. For each type of key features, there are multiple 
selection methods and encoding methods, resulting in a total of 13 types 
of images. CNN is trained on each kind of image, and the output scores of CNNs 
are fused into the final score for final recognition. 

\subsection{Feature extraction}
The spatial features studied in this paper include joint-joint distances, 
joint-joint vectors, joint-joint orientations, joint-line distances and 
line-line angles which were introduced in~\cite{Zhang2017}. In this paper, 
every action is assumed to be performed by two subjects, the main subject and 
an auxiliary other. In cases where there is only one person in the sequence, a 
'shadow subject' copied from main subject is adopted. Suppose each subject has 
$n$ joints, then in each frame there will be $N=2\times n$ joints. Let 
$p_j=(x,y,z),j\in N$ denote the 3D coordinate (Euclidean space) of the $j_{th}$ 
joint in a frame. The five features at frame $t$ are calculated as follows:
	
\begin{align}
	JJd^t_{jk}& = ||p^t_j-p^t_k|| \\
	JJv^t_{jk} & =  p^t_j-p^t_k \\
	JJo^t_{jk} & =  JJv^t_{jk}/JJd^t_{jk}\\
	JLd^t_{jkm} & =  JJv^t_{jk}\otimes  JJv^t_{jm}/JJd^t_{km}\\
	LLa^t_{jkmn} & = arccos(JJo^t_{jk}\odot JJo^t_{mn})
\end{align}
where $j,k,m,n\in N$ are the joint indices, $\otimes$ is cross 
product and $\odot$ is dot product. Meanwhile, $j \neq k$ in equations 
(1-4), $j\neq m\neq k$ in equation (4), and $(i,k)\neq (m,n)$ in equation 
(5). 
	
In total, there are $C^2_{50}=1225$ dimensions of the JJd feature, 
$3 \times 1225=3675$ dimensions of JJv and JJo features. There are also 1225 
lines, resulting in $1225 \times 48=58800$ dimensions of JLd feature and 
$C^2_{1225}=749700 $ dimensions of LLa feature. The resulting high dimensional 
feature space is neither cost-effective nor robust.
	
\subsection{Feature selection}
Feature selection is conducted by selecting key joints and key lines to reduce 
the number of combinations. The selection follows the principle that selected 
features should contain as much information as possible and be invariant to 
viewpoints and actions. Based on the observation that the motions are mainly 
located on the ends of skeletons and are usually locally sensitive, three 
strategies are proposed to select key joints for joint-joint feature 
calculation.

Joint strategy one (JS1): only the relations of joints within the same 
subject are considered, resulting in $2*C^2_{25}=600$ dimensional JJd feature. 
JS2: twelve joints from each subject are used, resulting in $C^2_{24}=276$ 
dimensional JJd feature. The joints start from 'middle of the spine' and are all 
two-steps away from the others.  JS3: eleven joints from each subject are used, 
resulting in $C^2_{22}=231$ dimensional JJd feature. The joints start from 
'base of the spine' and are all two-steps away from the others. 
	
	\begin{figure*}[!t]
		\centering
		\includegraphics[width=\linewidth]{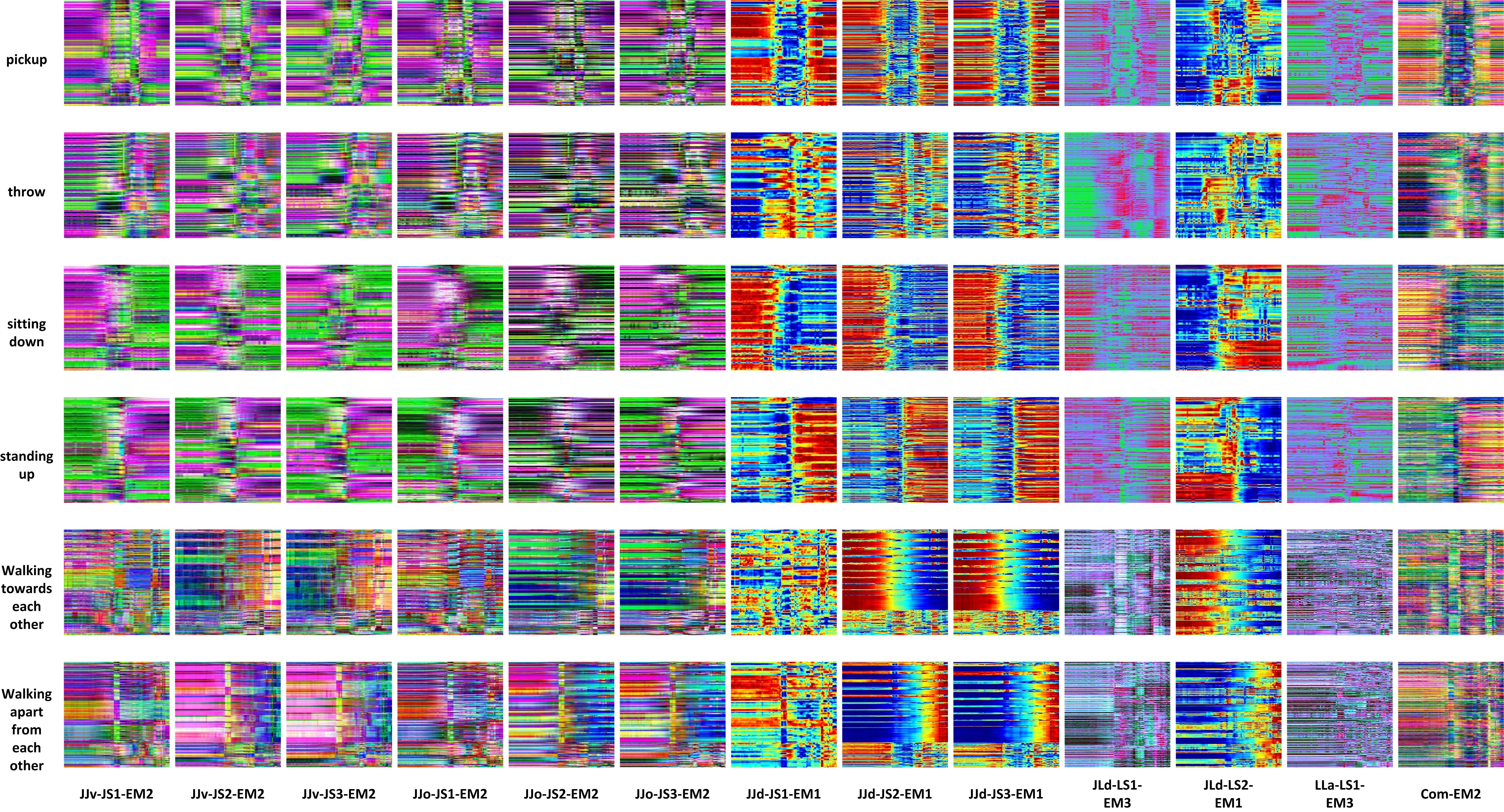}
		\caption{Samples generated by the proposed method on NTU RGB+D 
Dataset. Six samples from different actions are visualized. The images in each 
row are generated from the same sample, and the images in each column are 
generated using the same method. The images within the same row represent the 
difference of methods, and the images within the same column represent the 
difference between action classes.}
		\label{fig:samples}
	\end{figure*}
	
Two strategies are used to select key lines. Line strategy one (LS1): 
adopting the method in~\cite{Zhang2017} to select 39 lines from the main 
subject, resulting in $897$ dimensional JLd feature and $741$ dimensional LLa 
feature. LS2: using joints selected via JS3 to generate lines, and for each line 
the joints within two-step distance from end joints are used to calculate JLd 
feature, resulting in $ 570$ dimensional JLd feature.
	
\subsection{Color encoding}
Inspired by~\cite{Li2017}, color images are used to encode the spatial 
features to capture temporal information. Specifically, each column in the image 
represents spatial features in a frame, and each row represents the sequence of 
a specific feature. In this way, the textures represent the spatio-temporal 
information of skeleton sequences. Given a skeleton sequence with $T$ frames, 
$N$-dimensional features (scalar/vector) are extracted for each frame. 
The following three methods are used to encode the $N\times T$ feature into a 
$H(eight)\times W(idth)$ sized color image ($256\times256$ in this paper). 
      
      Encoding method one (EM1): for scalar features including JJd, JLd and LLa, 
the jet colorbar \cite{Li2017} is adopted to encode RGB channels jointly. The 
RGB value of pixel at $h_{th}$ row and $w_{th}$ column is 
      \begin{equation}
      \overline{RGB}(h,w) = colorbar((f^w_h-\min F_h)/(\max (F_h-\min F_h))
      \label{equ:colorbar}
      \end{equation}
      where $f^w_h$ is the value of the $h_{th}$ feature at $w_{th}$ frame, and 
$f^w_h = f^{T\times w/W}_{N\times h/H}$, i.e. the features are resized to $H*W$ 
using bilinear interpolation. $F_h=\{f^1_h,f^2_h,...,f^T_h\}$, $colorbar()$ is 
a mapping function which maps $[0,1]$ to corresponding RGB colors. 
      
      EM2: for vector features like JJo and JJv, RGB channels are encoded based 
on XYZ values respectively as follows:
      \begin{equation}
      \overline{RGB}(h,w) = (\bar{f}^w_h-\min \bar{F}_h)/(\max (\bar{F}_h-\min 
\bar{F}_h)
      \label{equ:xyz}
      \end{equation}
      where $f^w_h\in R^3$ is the vector of  $h_{th}$ feature at $w_{th}$ frame. 
Note that the operations are applied on each dimension.
      
      EM3: this method encodes RGB channels based on scalar features from both 
subjects. Specifically, red channel is encoded based on features of main 
subject, green channel is encoded using features of the auxiliary subject, and 
blue channel is encoded based on both features. The encoding method is 
formulated as follows:
      \begin{equation}
      \begin{aligned}
      R(h,w) & = 1 - (f^w_h-\min F_h)/(\max (F_h-\min F_h)\\
      G(h,w) & = (v^w_h-\min V_h)/(\max (V_h-\min V_h)\\
      B(h,w) & = 4\times R \times G
      \end{aligned}
      \label{equ:rgb}
      \end{equation}
      where $f,F$ and $v,V$ represent features from main subject and other 
subject specifically. 
      
      \subsection{CNN training and score fusion}
      In this paper, the Caffenet (a version of 
Alexnet~\cite{krizhevsky2012imagenet}) is adopted as the CNN model. The 
protocols used in~\cite{Li2017,Shahroudy2016} are adopted to train the CNN 
models from scratch. Given a testing skeleton sequence, thirteen types of images 
are generated and each type of image is recognized with a trained CNN model. All 
the outputs (scores) of the CNN models are then fused into a final score by 
element-wise multiplication, which has been verified 
in~\cite{Pichaocvpr2017,Li2017}. The fusion is done as follows:
      \begin{equation}
      label=F_{max}(v_1\circ v_2 \cdots v_{12} \circ v_{13})
      \end{equation}
      where $v$ are the score vectors, $\circ$ is the element-wise 
multiplication, and $F_{max}(\ldotp)$ is a function to find the index of the 
maximum element.
      
      \subsection{Implementation details}
      Joint coordinates are normalized in a way similar to the method 
in~\cite{Shahroudy2016}, where the spine lengths of the same subject in each 
frame (from `base of the spine' to `spine') are normalized to 1, and the other 
limb lengths are scaled in equal proportions. The scheme to select the main 
subject is adopted from~\cite{Shahroudy2016}, where the skeleton sequence having 
larger variations is set to be the main subject. Before selection, joint 
coordinates are translated from camera coordinate system to the body coordinate 
system, as described in \cite{Zhang2017}. The spatial features are directly 
calculated from normalized skeleton data to reduce the deviation introduced by 
the coordinate transformation.
      
      Caffe was adopted as the CNN platform and a Nvidia Titan X GPU was used to 
run the experiments. The CNNs were trained using stochastic gradient descent 
(SGD) for a total of 30000 iterations. The models were trained from scratch and 
the weights were initialized using Gaussian Filter. The multi-step scheme was 
used to train the CNNs with step sizes as 10000, 18000, 24000, 28000 
specifically. The learning rate was initially set to 0.01 and multiplied by 
$0.1$ every epoch.

      \section{Experiment Results}
      \label{sec:experiment}
      The proposed method was evaluated on NTU RGB+D Dataset. Currently, NTU 
RGB+D Dataset \cite{Shahroudy2016} is the largest dataset for action 
recognition. It has 56578 samples of 60 different actions classes, which are 
captured under 18 settings with different camera viewpoints and heights. The 
actions include single-subject cases and multi-subject interaction cases and are 
performed by 40 subjects aged between 10 and 35. This dataset is challenging and 
there are two types of protocols for evaluation of methods, cross-subject and 
cross-view. In this paper, the cross-view protocol is used. The effectiveness of
different types of spatial features, different joint selection schemes were 
evaluated. 
      
		\begin{table}[!t]
			\centering
			\caption{Evaluation results of different features and encoding methods. }
			\label{tab:result_all}
			\begin{tabular}{|c|c|c|c|c|}
				\hline Feature & Method & Accuracy &\multicolumn{2}{c|}{Fused Accuracy}\\
				\hline \multirow{3}{*}{JJv}& JS1-EM2&62.45\%&\multirow{3}{1cm}{75.23\%}& \multirow{13}{*}{82.31\%}\\
				\cline{2-3}& JS2-EM2 &65.12\% &&\\
				\cline{2-3}& JS3-EM2 &69.02\%&&\\
				\cline{1-4} \multirow{3}{*}{JJo}& JS1-EM2& 64.11\%&\multirow{3}{1cm}{73.51\%}&\\
				\cline{2-3}&JS2-EM2$^2$ & 55.14\%&&\\
				\cline{2-3}& JS3-EM2& 63.30\%&&\\
				\cline{1-4} \multirow{3}{*}{JJd}& JS1-EM1& 59.18\%&\multirow{3}{1cm}{73.01\%}&\\
				\cline{2-3}& JS2-EM1& 62.86\%&&\\
				\cline{2-3}& JS3-EM1& 62.95\%&&\\
				\cline{1-4} \multirow{2}{*}{JLd}& LS1-EM3& 63.08\%&\multirow{3}{1cm}{76.20\%}&\\
				\cline{2-3}& LS2-EM1& 59.71\% &&\\
				\cline{1-4} LLa & LS1-EM3& 62.57\%&62.57\%&\\
				\cline{1-4} Com$^1$ & JS1\&LS1-EM2 &62.00\% &62.00\%&\\
				\hline
			\end{tabular}
			{\flushleft \footnotesize  Note 1: this method encodes the RGB channels based on JJd, JLd and LLa respectively. 
				Note 2: this method is not used for final score fusion. }
		\end{table}	
		
			\begin{table}[!t]
				\centering
				\caption{Experimental results (accuracy) on NTU RGB+D Dataset}
				\label{tab:ntu_result}
				\begin{tabular}{|c|c|}
					\hline Method & Accuracy \\
					\hline Lie Group\cite{Vemulapalli2014} & 52.76\% \\
					Dynamic Skeletons\cite{Ohn-Bar2013} & 65.22\% \\
					HBRNN\cite{Du2015} & 63.97\% \\
					Deep RNN\cite{Shahroudy2016} & 64.09\% \\
					Part-aware LSTM\cite{Shahroudy2016} & 70.27\% \\
					ST-LSTM+Trust Gate\cite{Liu2016} & 77.70\% \\
					JTM\cite{Wang2016} & 75.20\% \\
					Geometric Features\cite{Zhang2017} & 82.39\% \\
					STA-LSTM\cite{Song2017}	& 81.20\%\\
					\hline
					Proposed Method & 82.31\% \\
					\hline
				\end{tabular}
			\end{table}
			
\subsection{Evaluation of spatial features}
The results of individual features and different encoding methods are 
 listed in Table~\ref{tab:result_all}, as well as results of 
score-multiplication fusion. There are five features evaluated, each of which 
was evaluated with different feature (joint) selection methods and different 
encoding methods. The methods are denoted in the form `feature selection method 
- encoding method', which have been described in 
Section~\ref{sec:method}.
	 
As illustrated in Fig.~\ref{fig:samples}, images generated from samples 
of different actions have discriminative textures. In addition, the spatial 
features are encoded into different textures by different methods. 
	
From Table~\ref{tab:result_all}, it can be seen that the JJv feature is the best 
joint-joint feature, based on the comparisons of single results and fused 
results. Moreover, JLd seems to be the best feature among the five types of 
features, which coincidences with the observations reported by~\cite{Zhang2017}. 
Among the three kinds of joint selection methods, JS3 generally works better 
than the other two. This observation suggests that some of the joints are 
noise with regard to this task, which is consistent with the above analysis.

From Table~\ref{tab:ntu_result} the results indicate that, 
compared with methods based hand-crafted features and those based on deep 
learning (RNNs and CNNs), the proposed method achieved state-of-the-art 
results. 
	
\section{Conclusions}
\label{sec:conclusion}
In this paper, a method for skeleton-based action recognition using CNNs is 
proposed. This method explored encoding different spatial features into texture 
color images and achieved state-of-the-art results on NTU RGB+D Dataset. The 
experimental results indicated the effectiveness of texture images when used as 
 spatio-temporal information representation, and the effectiveness of joint 
selection strategies for robust and cost-efficient computation. 
	
	% References should be produced using the bibtex program from suitable
	% BiBTeX files (here: strings, refs, manuals). The IEEEbib.bst bibliography
	% style file from IEEE produces unsorted bibliography list.
	% -------------------------------------------------------------------------


\begin{thebibliography}{10}

\bibitem{li2010action}
Wanqing Li, Zhengyou Zhang, and Zicheng Liu,
\newblock ``Action recognition based on a bag of {3D} points,''
\newblock in {\em Proc. IEEE Conference on CVPR Workshops (CVPRW)}, 2010, pp.
  9--14.

\bibitem{Wang2015}
Pichao Wang, Wanqing Li, Zhimin Gao, Chang Tang, Jing Zhang, and Philip
  Ogunbona,
\newblock ``Convnets-based action recognition from depth maps through virtual
  cameras and pseudocoloring,''
\newblock in {\em Proc. ACM Conference on Multimedia}, 2015, pp. 1119--1122.

\bibitem{pichaoTHMS}
Pichao Wang, Wanqing Li, Zhimin Gao, Jing Zhang, Chang Tang, and Phlip
  Ogunbona,
\newblock ``Action recognition from depth maps using deep convolutional neural
  networks,''
\newblock {\em IEEE Transactions on Human-Machine Systems}, vol. 46, no. 4, pp.
  498--509, 2016.

\bibitem{Pichaocvpr2017}
Pichao Wang, Wanqing Li, Zhimin Gao, Yuyao Zhang, Chang Tang, and Philip
  Ogunbona,
\newblock ``Scene flow to action map: A new representation for {RGB-D} based
  action recognition with convolutional neural networks,''
\newblock in {\em CVPR}, 2017.

\bibitem{Zhang2017}
Songyang Zhang, Xiaoming Liu, and Jun Xiao,
\newblock ``On geometric features for skeleton-based action recognition using
  multilayer {LSTM} networks,''
\newblock in {\em Proc. IEEE Winter Conference on Applications of Computer
  Vision (WACV)}, 2017.

\bibitem{xia2012view}
Lu~Xia, Chia-Chih Chen, and JK~Aggarwal,
\newblock ``View invariant human action recognition using histograms of {3D}
  joints,''
\newblock in {\em Proc. IEEE Conference on CVPR Workshops (CVPRW)}, 2012, pp.
  20--27.

\bibitem{wang2014mining}
Pichao Wang, Wanqing Li, Philip Ogunbona, Zhimin Gao, and Hanling Zhang,
\newblock ``Mining mid-level features for action recognition based on effective
  skeleton representation,''
\newblock in {\em Proc. IEEE Conference on Digital lmage Computing: Techniques
  and Applications (DlCTA)}, 2014, pp. 1--8.

\bibitem{Vemulapalli2014}
Raviteja Vemulapalli, Felipe Arrate, and Rama Chellappa,
\newblock ``Human action recognition by representing 3d skeletons as points in
  a lie group,''
\newblock in {\em CVPR}, 2014, pp. 588--595.

\bibitem{Wang2016}
Pichao Wang, Zhaoyang Li, Yonghong Hou, and Wanqing Li,
\newblock ``Action recognition based on joint trajectory maps using
  convolutional neural networks,''
\newblock in {\em Proc. ACM Conference on Multimedia}, 2016, pp. 102--106.

\bibitem{Hou2016}
Yonghong Hou, Zhaoyang Li, Pichao Wang, and Wanqing Li,
\newblock ``Skeleton optical spectra based action recognition using
  convolutional neural networks,''
\newblock {\em IEEE Transactions on Circuits and Systems for Video Technology},
  2016.

\bibitem{Li2017}
Chuankun Li, Yonghong Hou, Pichao Wang, and Wanqing Li,
\newblock ``Joint distance maps based action recognition with convolutional
  neural network,''
\newblock {\em IEEE Signal Processing Letters}, 2017.

\bibitem{Shahroudy2016}
Amir Shahroudy, Jun Liu, Tian-Tsong Ng, and Gang Wang,
\newblock ``{NTU RGB+D}: A large scale dataset for {3D} human activity
  analysis,''
\newblock in {\em CVPR}, June 2016.

\bibitem{krizhevsky2012imagenet}
Alex Krizhevsky, Ilya Sutskever, and Geoffrey~E Hinton,
\newblock ``Imagenet classification with deep convolutional neural networks,''
\newblock {\em Advances in Neural Information Processing Systems}, pp.
  1097--1105, 2012.

\bibitem{Ohn-Bar2013}
Eshed Ohn-Bar and Mohan Trivedi,
\newblock ``Joint angles similarities and hog2 for action recognition,''
\newblock in {\em Proc. IEEE Conference on CVPR Workshops (CVPRW)}, 2013, pp.
  465--470.

\bibitem{Du2015}
Yong Du, Wei Wang, and Liang Wang,
\newblock ``Hierarchical recurrent neural network for skeleton based action
  recognition,''
\newblock in {\em CVPR}, 2015, pp. 1110--1118.

\bibitem{Liu2016}
Jun Liu, Amir Shahroudy, Dong Xu, and Gang Wang,
\newblock ``Spatio-temporal lstm with trust gates for 3d human action
  recognition,''
\newblock in {\em ECCV}, 2016, pp. 816--833.

\bibitem{Song2017}
Sijie Song, Cuiling Lan, Junliang Xing, Wenjun Zeng, and Jiaying Liu,
\newblock ``An end-to-end spatio-temporal attention model for human action
  recognition from skeleton data,''
\newblock in {\em Proc. AAAI Conference on Artificial Intelligence}, 2017, pp.
  4263--4270.

\end{thebibliography}
\end{document}